\documentclass[11pt,a4paper]{article}
\usepackage[hyperref]{acl2019}
\usepackage{times}
\usepackage{latexsym}
\usepackage[pdftex]{graphicx}  
\usepackage{amsmath}
\usepackage{bm}
\usepackage{enumitem}
\setlist{itemsep=0px,parsep=0px,topsep=0.5\baselineskip,itemindent=-2mm}%,leftmargin=15px}
\usepackage{url}
\usepackage[farskip=0pt]{subfig}
\usepackage{lscape}
\usepackage[pass]{geometry}
\usepackage{float}

\newcommand{\bert}{\textsc{Bert}}
\newcommand{\mbert}{\mbox{\textsc{M}-\bert{}}}
\newcommand{\enbert}{\mbox{\textsc{En}-\bert{}}}
\newcommand{\pos}{\textsc{pos}}
\newcommand{\ner}{\textsc{ner}}

\aclfinalcopy % Uncomment this line for the final submission
\def\aclpaperid{2122} %  Enter the acl Paper ID here

\title{How multilingual is Multilingual BERT?}

\author{Telmo Pires\thanks{\ \ Google AI Resident.} \qquad Eva Schlinger \qquad Dan Garrette \\
  Google Research \\ 
  \texttt{\{telmop,eschling,dhgarrette\}@google.com}}

\date{}

\begin{document}
\maketitle

\begin{abstract}
In this paper, we show that Multilingual BERT (\mbert{}), released by \citet{devlin2018bert} as a single language model pre-trained from monolingual corpora in 104 languages, is surprisingly good at zero-shot cross-lingual model transfer, in which task-specific annotations in one language are used to fine-tune the model for evaluation in another language.
To understand why, we present a large number of probing experiments, showing that transfer is possible even to languages in different scripts, that transfer works best between typologically similar languages, that monolingual corpora can train models for code-switching, and that the model can find translation pairs.
From these results, we can conclude that \mbert{} does create multilingual representations, but that these representations exhibit systematic deficiencies affecting certain language pairs.
\end{abstract}

%\nocite{conneau2018cramsinglevector}
\section{Introduction}

% \dg{I can't decide whether this should be "Pre-training LMs enables generalization, we are interested in \emph{cross-lingual} generalization, we investigate using M-BERT because it's particularly well suited for zero-shot", or "Pre-training LMs enables generalization, M-BERT is an LM that can be used for zero-shot cross-lingual transfer, we investigate whether LMs are able to generalize across languages. "}

% \dg{Basic background on the topic.}
Deep, contextualized language models provide powerful, general-purpose linguistic representations that have enabled significant advances among a wide range of natural language processing tasks \cite{peter2018elmo,devlin2018bert}. These models can be pre-trained on large corpora of readily available unannotated text, and then fine-tuned for specific tasks on smaller amounts of supervised data, relying on the induced language model structure to facilitate generalization beyond the annotations.
Previous work on model probing has shown that these representations are able to encode, among other things, syntactic and named entity information, but they have heretofore focused on what models trained on English capture about English \cite{D18-1179,tenney2018what, tenney2019acl}.

% \dg{What are we doing and why are we doing it this way? Trying to avoid the reviewer asking "Why did you only try BERT?"}: I don't think Elmo, for example, ever released a multilingual checkpoint.
In this paper, we empirically investigate the degree to which these representations generalize \emph{across} languages.  We explore this question using Multilingual BERT (henceforth, \mbert{}), released by \citet{devlin2018bert} as a single language model pre-trained on the concatenation of monolingual Wikipedia corpora from 104 languages.\footnote{https://github.com/google-research/bert}  \mbert{} is particularly well suited to this probing study because it enables a very straightforward approach to zero-shot cross-lingual model transfer: we fine-tune the model using task-specific supervised training data from one language, and evaluate that task in a different language, thus allowing us to observe the ways in which the model generalizes information across languages.

Our results show that \mbert{} is able to perform cross-lingual generalization surprisingly well.
% , and does so despite the fact that unlike much of the existing work on cross-lingual transfer, it is not trained with an explicit objective encouraging translation-equivalent inputs to have similar representations \citep[\textit{inter alia}]{lample2019crosslingualpretraining,artetxe2018massively}.
More importantly, we present the results of a number of probing experiments designed to test various hypotheses about how the model is able to perform this transfer. Our experiments show that while high lexical overlap between languages improves transfer, \mbert{} is also able to transfer between languages written in different scripts---thus having \emph{zero} lexical overlap---indicating that it captures multilingual representations.  We further show that transfer works best for typologically similar languages, suggesting that while \mbert{}'s multilingual representation is able to map learned structures onto new vocabularies, it does not seem to learn systematic transformations of those structures to accommodate a target language with different word order.

% More importantly, we present the results of a large number of probing experiments designed to test various hypotheses about how the model is able to perform this transfer. For example, we investigate how much the model relies on lexical similarity by experimenting with Named Entity Recognition (\ner{}), a task that benefits from lexical overlap, and Part of Speech (\pos{}) tagging, which depends on learning latent structure.  We run experiments on a large number of language pairs, which allows us to see that, for example, transfer between superficially similar languages, like Spanish and Italian, works well, but so does transfer from Urdu to Hindi, a pair of languages that share a similar grammatical structure but have virtually \emph{zero} lexical overlap since they are written in different scripts.

% But why is it able to generalize to languages it didn't see (except during pre-training)? Is it because of different languages having common word pieces (word piece overlap)? Or memorizing context? Is model pre-training learning word order, independent of a specific language? What is the structure of the feature space?

% INCLUDE THIS? In this BERT+constituency paper \cite{kitaev2018multilingual}, they have an interesting piece of results showing that BERT pre-trained on Chinese does better than randomly initialized BERT on the English PTB task!

\section{Models and Data}

% \dg{Should we abbreviate "Multilingual BERT" as "M-BERT" or something for space? "MultiBERT"?}

Like the original English BERT model (henceforth, \enbert{}), \mbert{} is a 12 layer transformer
% , with 768 hidden units per layer
\citep{devlin2018bert}, but instead of being trained only on monolingual English data with an English-derived vocabulary, it is trained on the Wikipedia pages of 104 languages with a shared word piece vocabulary.
% (size 110k). 
It does not use any marker denoting the input language, and does not have any explicit mechanism to encourage translation-equivalent pairs to have similar representations.
% multilinguality, such as in \citet{lample2019crosslingualpretraining}: (CITE COMMON THINGS) no parallel data, as in \citet{artetxe2018massively}, no projection based on bilingual dictionaries, ... 
% In some of our experiments, we compare to the also use the English BERT checkpoint for comparison. It is a similar model, but trained only on English data and with an English specific word piece vocabulary.

For \ner{} and \pos{}, we use the same sequence tagging architecture as \citet{devlin2018bert}. We tokenize the input sentence, feed it to \textsc{Bert}, get the last layer's activations, and pass them through a final layer to make the tag predictions.
% For each word piece $w_i$, the probability distribution over output tags is:
% % \begin{align*}
% $p(\bm{y}_i \mid w_i) = \text{softmax}(W\cdot\bm{t}_i + b)$,
% % \end{align*}
% where $\bm{t}_i$ is the last layer's representation for word piece $w_i$.
The whole model is then fine-tuned to minimize the cross entropy loss for the task. When tokenization splits words into multiple pieces, we take the prediction for the first piece as the prediction for the word.

% \dg{Move hyperparameter settings to an appendix?}
% All models were fine-tuned with a batch size of $32$, and a maximum sequence length of $128$ for $3$ epochs. We use a learning rate of $3\mathrm{e}{-5}$ with learning rate warmup during the first $10\%$ of steps, and linear decay afterwards. We also apply $10\%$ dropout on the last layer.

\subsection{Named entity recognition experiments}

We perform \ner{} experiments on two datasets: the publicly available CoNLL-2002 and -2003 sets, containing Dutch, Spanish, English, and German \cite{sang2002conll,sang2003conll};
% , with labels Person, Location, Organization, and Miscellaneous.
% converted to use the BOISE sequence tagging scheme, which has been shown to improve model performance \cite{ratinov2009ner,dai2015ner}
and an in-house dataset with 16 languages,\footnote{Arabic, Bengali, Czech, German, English, Spanish, French, Hindi, Indonesian, Italian, Japanese, Korean, Portuguese, Russian, Turkish, and Chinese.} using the same CoNLL categories.
Table \ref{tab:conll} shows \mbert{} zero-shot performance on all language pairs in the CoNLL data.

\begin{table}[t!]
\centering
\small
\begin{tabular}{lllll}
  Fine-tuning \textbackslash{} Eval & \textsc{en} & \textsc{de} & \textsc{nl} & \textsc{es} \\
  \hline
  \textsc{en} & \textbf{90.70} & 69.74 & 77.36 & 73.59 \\
  \textsc{de} & 73.83 & \textbf{82.00} & 76.25 & 70.03 \\
  \textsc{nl} & 65.46 & 65.68 & \textbf{89.86} & 72.10 \\
  \textsc{es} & 65.38 & 59.40 & 64.39 & \textbf{87.18} \\
\end{tabular}
\caption{\textsc{Ner} F1 results on the CoNLL data.}
\label{tab:conll}
\end{table}

% Each row is a model fine-tuned on one language. This table shows some interesting patterns. For example, when transferring from English or German, it gets $70\%$ or higher in the other languages. The scores for Spanish are lower, which can probably be explained by the significantly smaller training set ($\sim$60\% of the size of the other datasets).

\subsection{Part of speech tagging experiments}

We perform \pos{} experiments using Universal Dependencies (UD) \cite{nivre2016universaldependencies} data for 41 languages.\footnote{Arabic, Bulgarian, Catalan, Czech, Danish, German, Greek, English, Spanish, Estonian, Basque, Persian, Finnish, French, Galician, Hebrew, Hindi, Croatian, Hungarian, Indonesian, Italian, Japanese, Korean, Latvian, Marathi, Dutch, Norwegian (Bokmaal and Nynorsk), Polish, Portuguese (European and Brazilian), Romanian, Russian, Slovak, Slovenian, Swedish, Tamil, Telugu, Turkish, Urdu, and Chinese.} We use the evaluation sets from \citet{zeman2017conll}.
Table \ref{tab:pos_european} shows \mbert{} zero-shot results for four European languages. We see that \mbert{} generalizes well across languages, achieving over $80\%$ accuracy for all pairs.

% We fine-tune models for $41$  languages\footnote{Arabic, Bulgarian, Catalan, Czech, Danish, German, Greek, English, Spanish, Estonian, Basque, Persian, Finnish, French, Galician, Hebrew, Hindi, Croatian, Hungarian, Indonesian, Italian, Japanese, Korean, Latvian, Marathi, Dutch, Norwegian (Bokmaal and Nynorsk), Polish, Portuguese (European and Brazilian), Romanian, Russian, Slovak, Slovenian, Swedish, Tamil, Telugu, Turkish, Urdu, and Chinese.}, and evaluate them on the others, covering all pairs. We use the evaluation sets from \citet{zeman2017conll}.

% Table \ref{tab:pos_european} summarizes the results on UD for European languages (see Appendix E for the full table). We see that BERT generalizes well across languages, achieving over $80\%$ accuracy for all pairs.
%\es{point to full comparison table in supplementary material}

\begin{table}[t!]
\small
\centering
\begin{tabular}{llllll}
  Fine-tuning \textbackslash{} Eval & \textsc{en} & \textsc{de} & \textsc{es} & \textsc{it} \\
  \hline
  \textsc{en} & \textbf{96.82} & 89.40 & 85.91 & 91.60 \\
  \textsc{de} & 83.99 & \textbf{93.99} & 86.32 & 88.39 \\
  \textsc{es} & 81.64 & 88.87 & \textbf{96.71} & 93.71 \\
  \textsc{it} & 86.79 & 87.82 & 91.28 & \textbf{98.11} \\
\end{tabular}
% \caption{\textsc{Pos} accuracy on the UD data for a subset of European languages.}
\caption{\textsc{Pos} accuracy on a subset of UD languages.}
\label{tab:pos_european}
\end{table}

\section{Vocabulary Memorization \label{sec:vocab_memorization}}
Because \mbert{} uses a single, multilingual vocabulary, one form of cross-lingual transfer occurs when word pieces present during fine-tuning also appear in the evaluation languages.  In this section, we present experiments probing \mbert{}'s dependence on this superficial form of generalization: How much does transferability depend on lexical overlap? And is transfer possible to languages written in different scripts (\emph{no} overlap)?

\subsection{Effect of vocabulary overlap}

If \mbert{}'s ability to generalize were mostly due to vocabulary memorization, we would expect zero-shot performance on \ner{} to be highly dependent on word piece overlap, since entities are often similar across languages.
To measure this effect, we compute $E_\textit{train}$ and $E_\textit{eval}$, the sets of word pieces used in entities in the training and evaluation datasets, respectively, and define overlap as the fraction of common word pieces used in the entities:
% $$\text{overlap} = \frac{|E_\textit{train} \cap E_\textit{eval}|}{|E_\textit{train} \cup E_\textit{eval}|}$$
$\textit{overlap} = |E_\textit{train} \cap E_\textit{eval}|~/~|E_\textit{train} \cup E_\textit{eval}|$.

\begin{figure}
    \centering
    \includegraphics[width=7cm]{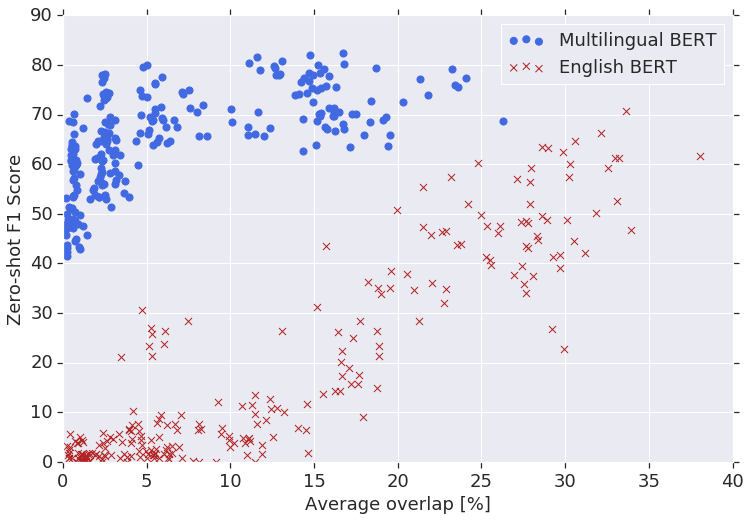}
    \caption{Zero-shot \ner{} F1 score versus entity word piece overlap among 16 languages.
    While performance using \enbert{} depends directly on word piece overlap, \mbert{}'s performance is largely independent of overlap, indicating that it learns multilingual representations deeper than simple vocabulary memorization.}
    \label{fig:ner_overlap}
\end{figure}

Figure \ref{fig:ner_overlap} plots \ner{} F1 score versus entity overlap for zero-shot transfer between every language pair in an in-house dataset of 16 languages, for both \mbert{} and \enbert{}.\footnote{Results on CoNLL data follow the same trends, but those trends are more apparent with 16 languages than with 4.}
% The points on the bottom left corner are for languages in a different script, so it should not come as a surprise that \enbert{} performs poorly.
% If word piece overlap was the main responsible for BERT's generalization, F1 score should go up as the overlap between entities increased. 
We can see that performance using \enbert{} depends directly on word piece overlap: the ability to transfer deteriorates as word piece overlap diminishes, and F1 scores are near zero for languages written in different scripts.
\mbert{}'s performance, on the other hand, is flat for a wide range of overlaps, and even for language pairs with almost no lexical overlap, scores vary between $40\%$ and $70\%$, showing that \mbert{}'s pretraining on multiple languages has enabled a representational capacity deeper than simple vocabulary memorization.\footnote{Individual language trends are similar to aggregate plots.}

To further verify that \enbert{}'s inability to generalize is due to its lack of a multilingual representation and not an inability of its English-specific word piece vocabulary to represent data in other languages, we evaluate on \emph{non}-cross-lingual \ner{} and see that it performs comparably to a previous state of the art model (see Table \ref{tab:conll_en}).

% It could be argued that the \enbert{} is unfairly disadvantaged due to having only English-derived word pieces. However, we believe this is not the reason it fails to generalize, as it performs comparably to previous state of the art models when tested on the fine-tuning language (see table \ref{tab:conll_en}).

\begin{table}[t!]
\centering
\small
\begin{tabular}{lllll}
  Model & \textsc{en} & \textsc{de} & \textsc{nl} & \textsc{es} \\
  \hline
  \citet{lample2016ner} & 90.94 & 78.76 & 81.74 & 85.75 \\
  \enbert{} & 91.07 & 73.32 & 84.23 & 81.84 \\
\end{tabular}
\caption{\ner{} F1 results fine-tuning and evaluating on the \emph{same} language (not zero-shot transfer).}
\label{tab:conll_en}
\end{table}

%To show that these results apply to all the languages, and not just to the aggregate (Simpson Paradox), we colour the dots by evaluation language in figure \ref{fig:ner_overlap_per_language}, and we see that the curves for different languages follow a similar shape.
% \begin{figure}
%     \centering
%     \includegraphics[width=7cm]{ner_overlap_per_language}
%     \caption{zero-shot F1 score vs entity overlap for the multilingual model. Same colour means same evaluation language. The overall pattern is the same for the individual languages.}
%     \label{fig:ner_overlap_per_language}
% \end{figure}

\subsection{Generalization across scripts}

\mbert{}'s ability to transfer between languages that are written in different scripts, and thus have effectively \emph{zero} lexical overlap, is surprising given that it was trained on separate monolingual corpora and not with a multilingual objective.  To probe deeper into how the model is able to perform this generalization, Table \ref{tab:pos_different_script} shows a sample of \pos{} results for transfer across scripts.
% (i.e., language pairs that hug the $y$-axis of Fig. \ref{fig:ner_overlap}). \tp{This may cause confusion, as the figure is for NER and the table for POS.}

Among the most surprising results, an \mbert{} model that has been fine-tuned using only \pos{}-labeled Urdu (written in Arabic script), achieves 91\% accuracy on Hindi (written in Devanagari script), even though it has never seen a single \pos{}-tagged Devanagari word. This provides clear evidence of \mbert{}'s multilingual representation ability, mapping structures onto new vocabularies based on a shared representation induced solely from monolingual language model training data.

However, cross-script transfer is less accurate for other pairs, such as English and Japanese, indicating that \mbert{}'s multilingual representation is not able to generalize equally well in all cases. A possible explanation for this, as we will see in section \ref{sec:typological}, is typological similarity. English and Japanese have a different order of subject, verb and object, while English and Bulgarian have the same, and \mbert{} may be having trouble generalizing across different orderings.

\begin{table}[t]
\small
\centering
\renewcommand{\arraystretch}{1} % Default value: 1
\subfloat{
\begin{tabular}{lll}
   & \textsc{hi} & \textsc{ur} \\
  \hline
  \textsc{hi} & \textbf{97.1} & 85.9 \\
  \textsc{ur} & 91.1 & \textbf{93.8} \\
  \\
\end{tabular}}
\hskip2.5em
%\quad
\subfloat{
\begin{tabular}{llll}
   & \textsc{en} & \textsc{bg} & \textsc{ja} \\
  \hline
  \textsc{en} & \textbf{96.8} & 87.1          & 49.4  \\
  \textsc{bg} & 82.2          & \textbf{98.9} & 51.6 \\
  \textsc{ja} & 57.4          & 67.2          & \textbf{96.5} \\
\end{tabular}}
\qquad
\caption{\pos{} accuracy on the UD test set for languages with different scripts. Row=fine-tuning, column=eval.\label{tab:pos_different_script}}
\end{table}

% Move this to conclusion?
%Regarding the generalization across scripts, we have an intuition as to why it is happening. It is known that word embedding spaces exhibit similar structure across languages \cite{mikilov2013embeddings}. Since in practice, all languages will share some word pieces (numbers, URLs, etc), which \emph{have} to be mapped to a shared space (because they are common), we hypothesize that this forces the word pieces that co-occur with them to also be mapped to a shared space. This effect then spreads to the other pieces, until different languages are close to a shared space. ALSO: EN TEXT IN OTHERS LANGUAGES.

\section{Encoding Linguistic Structure \label{sec:pos}}

In the previous section, we showed that \mbert{}'s ability to generalize cannot be attributed solely to vocabulary memorization, and that it must be learning a deeper multilingual representation.  In this section, we present probing experiments that investigate the nature of that representation: How does typological similarity affect \mbert{}'s ability to generalize? Can \mbert{} generalize from monolingual inputs to code-switching text?  Can the model generalize to transliterated text without transliterated language model pretraining?

% If \mbert{} is not just memorizing word pieces, is it capturing deeper structure? We now look at how linguistic typological similarity influences generalization, particularly the effect of word order, and evaluate \mbert{} on a code-switching dataset to see if it is able to generalize to linguistic structures unseen during pre-training.

%\paragraph{Does BERT generalize to languages that are syntactically different? What about different scripts?}
% \paragraph{Does BERT generalize because the test sets are ``easy''?} It is known that English POS taggers make ``egregious errors related to noun-verb ambiguity, despite having achieved $97\%+$ accuracy on the WSJ Penn Treebank'' \cite{ali2018nvdisambiguation}. We evaluate BERT fine-tuned on UD's languages on the Noun-Verb disambiguation challenge set from \citet{ali2018nvdisambiguation}.
%Is it just mapping the grammatical structure of the fine-tuning language to the target language?

\subsection{Effect of language similarity}
\begin{figure}
    \centering
    \includegraphics[width=7cm]{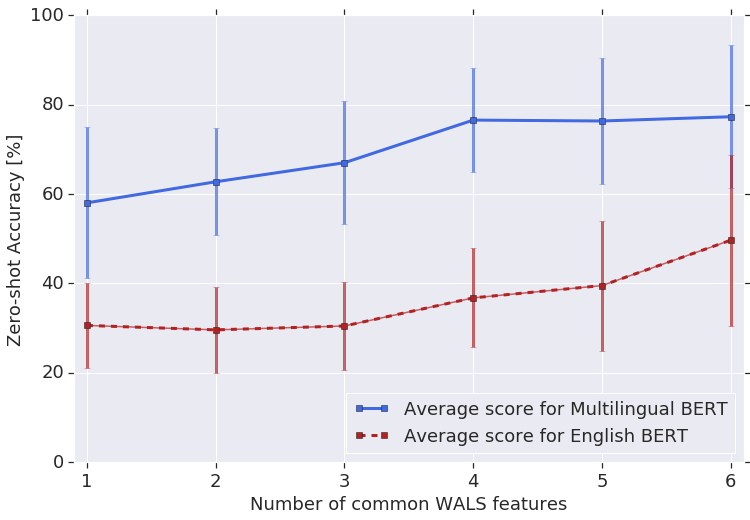}
    \caption{Zero-shot \pos{} accuracy versus number of common WALS features. Due to their scarcity, we exclude pairs with no common features.}
    \label{fig:wals}
\end{figure}

Following \citet{naseem2012selectivesharing}, we compare languages on a subset of the WALS features \cite{dryer2013wals} relevant to grammatical ordering.\footnote{81A (Order of Subject, Object and Verb), 85A (Order of Adposition and Noun), 86A (Order of Genitive and Noun), 87A (Order of Adjective and Noun), 88A (Order of Demonstrative and Noun), and 89A (Order of Numeral and Noun).}
Figure \ref{fig:wals} plots \pos{} zero-shot accuracy against the number of common WALS features. As expected, performance improves with similarity, showing that it is easier for \mbert{} to map linguistic structures when they are more similar, although it still does a decent job for low similarity languages when compared to \enbert{}.

% For comparison, we run the same experiments but, instead of fine-tuning \mbert{}, we extract the activations for all the 12 layers and train just a linear layer on top (red). We see that while the fine-tuning does improve performance, BERT by itself had already learned crosslingual POS features.

% In case we can't fit the plot, we have a table with the same information (except the error bars).
% \begin{table}[t!]
% \small
% \centering
% \begin{tabular}{l|llllll}
%   \# features & 1 & 2 & 3 & 4 & 5 & 6 \\
%   \hline
%   \mbert{} & 58.0 & 62.7 & 67.0 & 76.5 & 76.3 & 77.3 \\
%   \enbert{} & 30.5 & 29.6 & 30.4 & 36.7 & 39.5 & 49.6 \\
% \end{tabular}
% \caption{Zero-shot POS accuracy versus number of common WALS features. Due to their scarcity, we exclude pairs with no common features.\label{tab:wals}}
% \end{table}

\subsection{Generalizing across typological features
\label{sec:typological}}
%We now take a closer look to how \mbert{} performs when transferring between languages with different orders of Subject, Verb, and Object, and between languages with different orders of adjective and noun.
%For the latter, we search the evaluation datasets for instances of noun adjective in evaluation languages with this ordering, and look at the errors the models fine-tuned in adjective-noun languages make on these cases, and repeat a similar experiment on the the reverse direction.

% TELMO: is this footnote really necessary? I'm removing it.
%\footnote{I.e., the average score of the languages in each group.}
%\tp{For comparison fairness, we only include zero-shot pairs that are in different scripts?}
Table \ref{tab:wals} shows macro-averaged \pos{} accuracies for transfer between languages grouped according to two typological features: subject/object/verb order, and adjective/noun order\footnote{\textbf{SVO languages}: Bulgarian, Catalan, Czech, Danish, English, Spanish, Estonian, Finnish, French, Galician, Hebrew, Croatian, Indonesian, Italian, Latvian, Norwegian (Bokmaal and Nynorsk), Polish, Portuguese (European and Brazilian), Romanian, Russian, Slovak, Slovenian, Swedish, and Chinese. \textbf{SOV Languages}: Basque, Farsi, Hindi, Japanese, Korean, Marathi, Tamil, Telugu, Turkish, and Urdu.} \cite{dryer2013wals}. The results reported include only zero-shot transfer, i.e. they do not include cases training and testing on the same language.
% the score for transferring between a ``Subject, Verb, Object'' (SVO) language and a ``Subject, Object, Verb'' (SOV) one, averaged for all languages in each category. 
% SVO to SVO transfer is significantly better than SVO to SOV, indicating that \mbert{} might be having problems mapping between grammatical orders.
We can see that performance is best when transferring between languages that share word order features, suggesting that while \mbert{}'s multilingual representation is able to map learned structures onto new vocabularies, it does not seem to learn systematic transformations of those structures to accommodate a target language with different word order.

% "between groups" - this could be phrased better.
% Similarly, table \ref{tab:pos_an} reports the average results for transferring between a language where adjectives precede nouns (AN language) and one where nouns precede adjectives (NA language). The differences between groups are small, indicating that \mbert{} is doing a good job mapping between orders.

\begin{table}[t]
\small
\centering
\subfloat[Subj./verb/obj. order.\label{tab:pos_svo}]{
\begin{tabular}{lll}
   & SVO & SOV \\
  \hline
  SVO & \textbf{81.55} &         66.52  \\
  SOV &         63.98  & \textbf{64.22} \\
\end{tabular}}
\qquad
\subfloat[Adjective/noun order.\label{tab:pos_an}]{
\begin{tabular}{lll}
   & AN & NA \\
  \hline
  AN & \textbf{73.29} & 70.94 \\
  NA & 75.10 & \textbf{79.64} \\
\end{tabular}}
\qquad
\caption{Macro-average \pos{} accuracies when transferring between SVO/SOV languages or AN/NA languages. Row = fine-tuning, column = evaluation.}
\label{tab:wals}
\end{table}

\subsection{Code switching and transliteration}
Code-switching (CS)---the mixing of multiple languages within a single utterance---and transliteration---writing that is not in the language's standard script---present unique test cases for \mbert{}, which is pre-trained on monolingual, standard-script corpora.
% presents a unique problem for \nlp{} systems: it is a frequent phenomenon of informal communication, but annotated code-switched corpora are almost non-existent.  
Generalizing to code-switching is similar to other cross-lingual transfer scenarios, but would benefit to an even larger degree from a shared multilingual representation.  Likewise, generalizing to transliterated text is similar to other cross-script transfer experiments, but has the additional caveat that \mbert{} was not pre-trained on text that looks like the target.

% It is frequently accompanied by transliteration, which happens when text in one language is written in a different script (e.g., Hindi written in Latin-script).

We test \mbert{} on the CS Hindi/English UD corpus from \citet{bhat2018udcs}, which provides texts in two formats: \emph{transliterated}, where Hindi words are written in Latin script, and \emph{corrected}, where annotators have converted them back to Devanagari script.
Table \ref{tab:pos_cs} shows the results for models fine-tuned using a combination of monolingual Hindi and English, and using the CS training set (both fine-tuning on the script-corrected version of the corpus as well as the transliterated version).

% \begin{table}[t]
% \small
% \centering
% \begin{tabular}{lrr}
%   Fine-tuning set & Corrected & Transliterated \\
%   \hline
% %   \textsc{en} & 70.29 & 50.68 \\
% %   \textsc{hi} & 77.88 & 43.79 \\
%   Monolingual \textsc{hi} + \textsc{en} & 86.59 & 50.41 \\
%   \citet{ball2018codeswitching} & --- & 77.40\\
%   Code-switched \textsc{hi}/\textsc{en} & 90.56 & 85.64 \\
%   \citet{bhat2018udcs} & --- & 90.53 \\
% \end{tabular}
% \caption{\pos{} accuracy on the code-switched Hindi/English dataset from \citet{bhat2018udcs}.}
% \label{tab:pos_cs}
% \end{table}

\begin{table}[t]
\small
\centering
\begin{tabular}{lrr}
  & Corrected & Transliterated \\
  \hline
  \multicolumn{3}{l}{Train on monolingual \textsc{hi}+\textsc{en}} \\
  \quad{} \mbert{} & 86.59 & 50.41 \\
  \quad{} \citet{ball2018codeswitching} & --- & 77.40\vspace{0.5mm}\\ 
  \multicolumn{3}{l}{Train on code-switched \textsc{hi}/\textsc{en}}  \\
  \quad{} \mbert{} & 90.56 & 85.64 \\
  \quad{} \citet{bhat2018udcs} & --- & 90.53 \\
\end{tabular}
\caption{\mbert{}'s \pos{} accuracy on the code-switched Hindi/English dataset from \citet{bhat2018udcs}, on script-corrected and original (transliterated) tokens, and comparisons to existing work on code-switch \pos{}.}
\label{tab:pos_cs}
\end{table}

For script-corrected inputs, i.e., when Hindi is written in Devanagari, \mbert{}'s performance when trained only on monolingual corpora is comparable to performance when training on code-switched data, and it is likely that some of the remaining difference is due to domain mismatch.  This provides further evidence that \mbert{} uses a representation that is able to incorporate information from multiple languages.  

However, \mbert{} is not able to effectively transfer to a transliterated target, suggesting that it is the language model pre-training on a particular language that allows transfer to that language.  \mbert{} is outperformed by previous work in both the monolingual-only and code-switched supervision scenarios.  Neither \citet{ball2018codeswitching} nor \citet{bhat2018udcs} use contextualized word embeddings, but both incorporate explicit transliteration signals into their approaches.

% \mbert{} performs poorly on transliterated text, on which it was not pre-trained, indicating that it is during this step that it learns a shared representation. On the manually corrected script, on the other hand, it achieves much higher scores, showing us that it can generalize to mixed grammars.

%\mbert{} performs poorly on the ``raw'' text n which it was not pre-trained. This shows that it is in the pre-training step that it is learning the multilingual representation, as it can transfer across scripts on which it was pre-trained (see \textsc{hi}-\textsc{ur}). On manually corrected scripts (Hindi written in Devanagari, and English written in Latin-script), \mbert{} achieves a high score in a code-switched setting, giving further weight to the hypothesis that the representations are multilingual.

\section{Multilingual characterization  of the feature space \label{sec:vector_translation}}
In this section, we study the structure of \mbert{}'s feature space. If it is multilingual, then the transformation mapping between the same sentence in $2$ languages should not depend on the sentence itself, just on the language pair.

\subsection{Experimental Setup}
We sample $5000$ pairs of sentences from WMT16 \cite{wmt2016sharedtask} and feed each sentence (separately) to \mbert{} with no fine-tuning. We then extract the hidden feature activations at each layer for each of the sentences, and average the representations for the input tokens except \textsc{[cls]} and \textsc{[sep]}, to get a vector for each sentence, at each layer $l$, $v_\textsc{lang}^{(l)}$.
For each pair of sentences, e.g. $(v_{\textsc{en}_i}^{(l)}, v_{\textsc{de}_i}^{(l)})$, we compute the vector pointing from one to the other and average it over all pairs: $\bar{v}_{\textsc{en}\rightarrow \textsc{de}}^{(l)} = \frac{1}{M}\sum_i \left(v_{\textsc{de}_i}^{(l)} - v_{\textsc{en}_i}^{(l)}\right)$, where $M$ is the number of pairs.
Finally, we translate each sentence, $v_{\textsc{en}_i}^{(l)}$, by $\bar{v}_{\textsc{en}\rightarrow \textsc{de}}^{(l)}$, find the closest German sentence vector\footnote{In terms of $\ell_2$ distance.}, and measure the fraction of times the nearest neighbour is the correct pair, which we call the ``nearest neighbor accuracy''.
%\footnote{\textsc{[cls]} was trained for next sentence prediction, and not to be a representation of the input sentence, so it doesn't make sense to use it.}

%For each pair of parallel sentences, $(v_{\textsc{en}_i}^{(l)}, v_{\textsc{de}_i}^{(l)})$, we compute the vector pointing from one to the other: $v_{\textsc{en}\rightarrow \textsc{de}_i}^{(l)} = v_{\textsc{de}_i}^{(l)} - v_{\textsc{en}_i}^{(l)}$, and the average it over all pairs: $\bar{v}_{\textsc{en}\rightarrow \textsc{de}}^{(l)} = \frac{1}{M}\sum_i v_{\textsc{en}\rightarrow \textsc{de}_i}^{(l)}$, where $M$ is the number of pairs.

\subsection{Results}
\begin{figure}
    \centering
    \includegraphics[width=7cm]{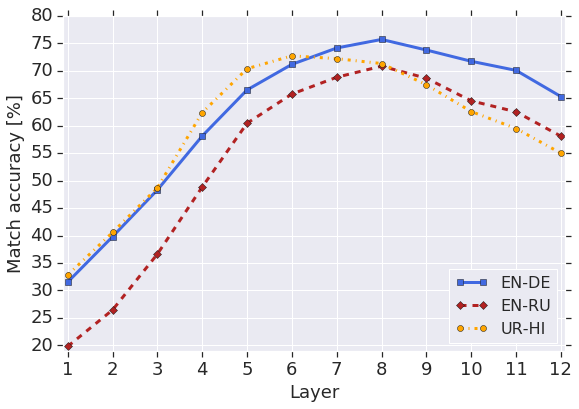}
    \caption{Accuracy of nearest neighbor translation for \textsc{en}-\textsc{de}, \textsc{en}-\textsc{ru}, and \textsc{hi}-\textsc{ur}.}
    \label{fig:vector_translation}
\end{figure}

In Figure \ref{fig:vector_translation}, we plot the nearest neighbor accuracy for \textsc{en}-\textsc{de} (solid line). It achieves over $50\%$ accuracy for all but the bottom layers,\footnote{Our intuition is that the lower layers have more ``token level'' information, which is more language dependent, particularly for languages that share few word pieces.} which seems to imply that the hidden representations, although separated in space, share a common subspace that represents useful linguistic information, in a language-agnostic way. Similar curves are obtained for \textsc{en}-\textsc{ru}, and \textsc{ur}-\textsc{hi} (in-house dataset), showing this works for multiple languages.

As to the reason why the accuracy goes down in the last few layers, one possible explanation is that since the model was pre-trained for language modeling, it might need more language-specific information to correctly predict the missing word.

\section{Conclusion}
% People know this by now.
%In this work, we performed an empirical analysis of \mbert{}. We showed that it performs much better on cross-lingual zero-shot learning tasks than anticipated, suggesting that it is learning a shared multilingual representation.

%Despite the surprising results \mbert{} achieves in zero-shot learning, 

In this work, we showed that \mbert{}'s robust, often surprising, ability to generalize cross-lingually is underpinned by a multilingual representation, without being explicitly trained for it.  The model handles transfer across scripts and to code-switching fairly well, but effective transfer to typologically divergent and transliterated targets will likely require the model to incorporate an explicit multilingual training objective, such as that used by \citet{lample2019crosslingualpretraining} or \citet{artetxe2018massively}.
% We believe that, in addition to shedding light on what large pretrained language models can do in terms of cross-lingual transfer, this work has analyzed where these models currently fail, indicating interesting directions for future research.
% The model naturally generalizes well Despite this, we believe there is room for improvement. 
% There have been some promising results from \citet{lample2019crosslingualpretraining}, and we believe that explicitly forcing multilingual representations will improve performance, particularly on pairs on which we have shown it doesn't work as well, such as English and Japanese.

As to why \mbert{} generalizes across languages, we hypothesize that having word pieces used in all languages (numbers, URLs, etc) which have to be mapped to a shared space forces the co-occurring pieces to also be mapped to a shared space, thus spreading the effect to other word pieces, until different languages are close to a shared space.

It is our hope that these kinds of probing experiments will help steer researchers toward the most promising lines of inquiry by encouraging them to focus on the places where current contextualized word representation approaches fall short.

%\cite{W18-3024}
%\cite{P18-1027}

\section{Acknowledgements}
We would like to thank Mark Omernick, Livio Baldini Soares,
% for providing the numbers for BERT fine-tuned on the Noun-Verb disambiguation training set, and 
%  for the help with the architectural challenges. We would also like to thank 
Emily Pitler, Jason Riesa, and Slav Petrov for the valuable discussions and feedback.

\bibliography{acl2019}
\bibliographystyle{acl_natbib}

\appendix

\section{Model Parameters}

All models were fine-tuned with a batch size of $32$, and a maximum sequence length of $128$ for $3$ epochs. We used a learning rate of $3\mathrm{e}{-5}$ with learning rate warmup during the first $10\%$ of steps, and linear decay afterwards. We also applied $10\%$ dropout on the last layer. No parameter tuning was performed.
We used the \texttt{BERT-Base, Multilingual Cased} checkpoint from \url{https://github.com/google-research/bert}.

\section{CoNLL Results for \enbert{}}

\begin{table}[H]
\centering
\small
\begin{tabular}{lllll}
  Fine-tuning \textbackslash Eval & \textsc{en} & \textsc{de} & \textsc{nl} & \textsc{es} \\
  \hline
  \textsc{en} & \textbf{91.07} & 24.38 & 40.62 & 49.99 \\
  \textsc{de} & 55.36 & \textbf{73.32} & 54.84 & 50.80 \\
  \textsc{nl} & 59.36 & 27.57 & \textbf{84.23} & 53.15 \\
  \textsc{es} & 55.09 & 26.13 & 48.75 & \textbf{81.84} \\
\end{tabular}
\caption{\ner{} results on the CoNLL test sets for \enbert{}. The row is the fine-tuning language, the column the evaluation language. There is a big gap between this model's zero-shot performance and \mbert{}'s, showing that the pre-training is helping in cross-lingual transfer.}
\end{table}

\section{Some \pos{} Results for \enbert{}}

\begin{table}[H]
\small
\centering
\begin{tabular}{llllll}
  Fine-tuning \textbackslash Eval & \textsc{en} & \textsc{de} & \textsc{es} & \textsc{it} \\
  \hline
  \textsc{en} & \textbf{96.94} & 38.31 & 50.38 & 46.07 \\
  \textsc{de} & 28.62 & \textbf{92.63} & 30.23 & 25.59 \\
  \textsc{es} & 28.78 & 46.15 & \textbf{94.36} & 71.50 \\
  \textsc{it} & 52.48 & 48.08 & 76.51 & \textbf{96.41} \\
\end{tabular}
\caption{\pos{} accuracy on the UD test sets for a subset of European languages using \enbert{}. The row specifies a fine-tuning language, the column the evaluation language. There is a big gap between this model's zero-shot performance and \mbert{}'s, showing the pre-training is helping learn a useful cross-lingual representation for grammar.}
\end{table}

\end{document}